\documentclass{ecai} 

\usepackage{latexsym}
\usepackage{amssymb}
\usepackage{amsmath}
\usepackage{amsthm}
\usepackage{booktabs}
\usepackage{enumitem}
\usepackage{graphicx}
\usepackage{color}
\usepackage{xcolor}
\usepackage{hyperref} 
\usepackage{adjustbox} 
\usepackage{makecell}  

\newcommand{\BibTeX}{B\kern-.05em{\sc i\kern-.025em b}\kern-.08em\TeX}

\begin{document}


\begin{frontmatter}

\paperid{0331} 

\title{A comparison of pipelines for the translation of a low resource language based on transformers}



\author[A,D]{\fnms{Chiara}~\snm{Bonfanti}\orcid{0009-0007-8015-7786}\thanks{Corresponding Author. Email: chiara.bonfanti@polito.it.}}
\author[A]{\fnms{Michele}~\snm{Colombino}\orcid{0009-0007-3248-1661}}
\author[A]{\fnms{Giulia}~\snm{Coucourde}\orcid{0009-0000-9051-386X}} 
\author[A,C]{\fnms{Faeze}~\snm{Memari}\orcid{0009-0003-3778-3619}} 
\author[A,B]{\fnms{Stefano}~\snm{Pinardi}\orcid{0000-0002-9639-1805}} 
\author[A]{\fnms{Rosa}~\snm{Meo}\orcid{0000-0002-0434-4850}} 

\address[A]{Department of Computer Science, University of Turin}
\address[B]{Department of Languages Studies, University of Turin}
\address[C]{Department of Humanistic Studies, University of Turin}
\address[D]{Department of Control and Computer Engineering, Politecnico di Torino}


\begin{abstract} 
This work proposes a comparison between three pipelines for training neural network architectures based on transformers to produce machine translators for a resource-poor language such as Bambara, a Mandè language spoken in Africa by about 14,188,850 people. The first pipeline involves training a simple transformer neural network to translate sentences from French into Bambara. The second pipeline is based on the fine-tuning of LLama3 (3B-8B) instructor models to use decoder-only architectures to generate translations from French to Bambara. The models of the first two pipelines were trained with different combinations of training hyperparameters to improve BLEU and chrF scores. These scores were evaluated both on the test sentences of each dataset and on official Bambara language benchmarks. The third pipeline uses language distillation with a student-teacher dual neural network architecture to integrate Bambara into a pre-trained LaBSE model. LaBSE models are agnostic, i.e. they contain a knowledge representation based on embeddings that are not language-dependent. The BERT extension technology is then applied to LaBSE to generate translations in the decoder. All pipelines have been tested on some common datasets: Dokotoro (medical domain) and Bayelemagaba (mixed domains). Experimental results show that the first pipeline, although simpler, achieves the best accuracy scores for translation ($10\%$ of BLEU, $21\%$ of chrF on the Bayelemagaba benchmark). These figures are in line with the results for low-resource language translation. On another dataset, Yiri, created specifically for this work as an integration from different sources, the first pipeline achieves a BLEU of $33.81\%$ and a chrF of $41\%$. The instructor-based approach shows better performance when applied to single datasets as opposed to aggregated multi-dataset collections. This suggests that the method is more effective at capturing dataset-specific patterns and nuances.
\end{abstract}

\end{frontmatter}


\newcommand{\pinardi}[1]{{\color{green} #1}}
\newcommand{\rosa}[1]{{\color{green} #1}}
\newcommand{\fm}[1]{{\color{green} #1}}
\newcommand{\mc}[1]{{\color{blue} #1}}
\newcommand{\cb}[1]{{\color{red} #1}}
\newcommand{\gc}[1]{{\color{orange} #1}}


\section{Introduction} 


The Yiri project is a project for social good to support low-resource languages (LRL), with a focus on African languages. Through the fine-tuning of large-language models on multilingual datasets, the use of specific corpora, and transfer learning techniques, the project seeks to improve semantic understanding, machine translation, and access to language technologies for communities underrepresented in the digital landscape. 
The study focuses on an underrepresented language of West Africa, Bambara, part of the Mandè family and spoken by approximately 14 million people. Specifically in the context of agricultural management, we are adopting a living lab approach to foster an open and collaborative environment to build capacity, promote knowledge sharing, and collaboration among individuals. In the short term, it is crucial to support communities to access some vital information, medical and agricultural, taking into account the specificity of the context. It is well known that the scarcity of textual data in LRLs hinders the development of machine translation and other language technologies. In this study, using bilingual corpora and experimenting with agricultural and medical domains, we aim to explore cross-domain strategies to improve access to information in local languages. Agriculture often represents the primary - or even sole -  source of income. In the long term, the Yiri project aims to empower local communities by supporting African education and cultural representation. 

This work builds upon these motivations, focusing on the development of Natural Language Processing (NLP) methods tailored for Bambara. While grounded in the social need to enhance information access in agriculture and healthcare, our approach also addresses key technical challenges in multilingual representation and model adaptation.

The main contributions of our study are as follows:
\begin{itemize}
    \item \textbf{Multi-domain bilingual dataset creation:} We constructed and integrated French–Bambara corpora covering agriculture and medicine, providing a new resource for cross-domain experimentation, named Yiri.
   
    \item \textbf{Novel distillation-based pipeline:} We designed LoReB, a new fine-tuning pipeline that leverages multilingual BERT models with knowledge distillation, for a low-resource language (Bambara).
   
   
    \item \textbf{Comparative evaluation of models:} We compared LoReB with an agnostic encoder equipped with an off-the-shelf translator in a pipeline that we named {\em Transformer From Scratch}. 
    The former two pipelines are compared with a third one, based on LLama3 (3B-8B) that we named {\em Instructor-based LLama3}. It is an instance of a causal language model driven by prompt instructions and based on the computation of query, keys, and values of self-attention in an approach with grouped multi-queries with keys and values in cache. This choice is justified by the need to highlight the trade-offs between different approaches in performance and abilities in generalization.
   
   
    
\end{itemize}

These contributions lay the groundwork for the methodological and technical developments discussed in the next sections.

\rosa{
} 


\section{Literature review}
\label{lit_rev}
In addition to the moral and humanistic motivations implicit in the ability to give voice to the unheard, this work is driven by the computational challenge posed by the use of LRLs in NLP.\\
There are two main hurdles to overcome: the lack of data and the optimization of computational resources.
At the time of this study, there are approximately 7000 languages spoken worldwide, many of which suffer from a lack of resources for training Large Language Models (LLMs) for many tasks~\cite{haddow-etal-2022-survey}, i.e. Neural Machine Translation (NMT), an advanced approach to automated translation.

\subsection{A computational point of view}
Recent research in computational linguistics has focused on Large Language Models (LLMs) for various NLP tasks, including language translation~\cite{LLMSurvey}, information retrieval and question answering. However, the presence of large amounts of data is a crucial prerequisite for the training of ML models. 
In the past, one possible approach to the significant obstacle of lack of documentation is to try to encapsulate the necessary knowledge for inference systems in a static Knowledge Base, represented by rule-based systems~\cite{pirinen-2019-building,sahala-etal-2020-babyfst}. Neural networks are also useful for their intrinsic ability to recognize patterns and features in documents, as it is as demonstrated in this work. Furthermore, it is possible to improve the management of computational resources with different methodologies through Adapters~\cite{Leong2023},  a technology used to overcome the limitations of data and computational resources. 

\subsubsection{Common ground: the use of resource-poor languages in NLP}
At the time of writing, there is no common definition in the literature other than the intuitive one: it has fewer resources than most common languages, such as English or French.
It is debated, as in ~\cite{popel2020transforming}, whether developed translation systems can qualitatively surpass human translation. Typically, to achieve these goals, one needs millions of pairs in the source and target languages.\\
This requirement, by definition, excludes languages such as Bambara, leaving a large research space unexplored. There is often a correlation between the number of speakers and the resources available, but there are important exceptions where widely spoken languages have few parallel data sets or languages. As can be seen in Table~\ref{tab:lang_pair}, languages such as Myanmar and Fon, which differ in the number of speakers by several orders of magnitude, can be considered low-resource languages from a computational point of view.
Lacking formal computational classification on what is a low-resource language, we follow the categorization proposed in~\cite{muller2021when} of Hard, Intermediate, and Easy languages. 
Bambara, which qualifies in between Hard and Medium language, is in a privileged position because it is UTF-8 supported and most of it is written in an alphabet based on common Latin with the addition of symbols from the International Phonetic Alphabet. 

\begin{table}[h]
\caption{Some language pairs and their available resources (from~\cite{haddow-etal-2022-survey}).\label{tab:lang_pair}}
\centering
\vspace{0.5cm}
\begin{tabular}{lrr}
\toprule 
\textbf{Language Pair} & \textbf{Speakers (approx.)} & \textbf{Parallel Phrases} \\ \hline
English–French     & 267M  & 280M     \\
English–Myanmar    & 30M   & 0.7M     \\
English–Fon        & 2M    & 0.035M   \\
French–Bambara     & 15M   & 0.3087M  \\
English–Bambara    & 15M   & 0.1824M  \\ \bottomrule 
\end{tabular}
\end{table}

\subsection{NLP background and techniques}
In NLP, growing attention is given to LLMs, enhanced by the debut of Chat-GPT in 2022.  Other companies, following Open-AI footsteps, released similar models such as  Bard~\cite{google2023} from Google or the LLaMA model family~\cite{touvron2023llama} by Meta. 
Important predecessors are N-Grams~\cite{shannon1948mathematical}, Recurrent Neural Networks (RNN)~\cite{rumelhart1986learning} and Long Short-Term Memory Neural Networks (LSTM)~\cite{hochreiter1997long}. The N-Grams involve a sequence of n-words, where each word likelihood depends on the previous one. Traditional neural networks enhance this representation by basing their learning model on the information available at a specific point in time, rather than on previous knowledge. However, this has the downside of memory limitations. It is addressed by RNNs incorporating internal loops. Despite this, RNNs still face challenges such as vanishing gradients and difficulties with long-term dependencies. LSTM modifies RNNs to effectively capture long-term dependencies and mitigate the vanishing gradient problems. 
The aforementioned models, in particular GPT, had a big impact on research worldwide
~\cite{picazo-sanchez2024}.
Transformers~\cite{vaswani2017attention} and specifically BERT models~\cite{BERT}, are recent architectures at the base of many LLMs: they incorporate self-attention mechanisms for context-aware word representations, enabling better performance across diverse NLP tasks. Their multi-head attention mechanism, coupled with deep feedforward networks and normalization techniques, facilitates efficient learning of complex linguistic patterns. 
Multilingual research in the field of NLP helps to tackle different challenges posed by different language structures. It also improves the content accessibility~\cite{doddapaneni2021primer}.\par{Bert enhanced NMT}, is a term used through this work concerning the fine-tuning of models related to the BERT family for neural machine translation. 
Usually, BERT is typically employed for fine-tuning rather than as a contextual embedding for downstream language understanding tasks, which could follow in the pipeline of text processing.
Preliminary investigations into the use of BERT in NMT indicate that utilising BERT as a contextual embedding yields superior outcomes compared to its application for fine-tuning ~\cite{zhu2020incorporatingbertneuralmachine}. This finding motivates the exploration of improved methods for utilising BERT in NMT. 
Such research literature is prolific, stemming works such as BERT-enhancement ~\cite{zhang2020bertjamboostingbertenhancedneural}, BERT-fused ~\cite{zhu2020incorporatingbertneuralmachine} and BERT-JAM ~\cite{ZHANG202184}.
All of them converge on the notion of how BERT-based representations are subsequently integrated with different mechanisms of the downstream LLM model. 
In this work, we proposed a novel algorithm, LoReB, designated as a kind of BERT-fused NMT, wherein BERT was initially employed only to extract representations for an input sequence. 

Of particular note is the mBART model ~\cite{chipman2021mbartmultidimensionalmonotonebart}, which has demonstrated effectiveness even in low-resource language environments ~\cite{tang2020multilingualtranslationextensiblemultilingual}. It advances multilingual sequence-to-sequence pre-training and enables cross-lingual generation with remarkable fluency and flexibility.
Complementary to this direction, Glot500 ~\cite{imanigooghari-etal-2023-glot500} is pushing the boundaries of language coverage by pre-training models in more than 500 languages, the majority of them resource-poor, emphasising inclusivity and linguistic diversity in multilingual NLP.

\subsubsection{Agnostic language models}
\label{subsubs:agnl}
Low-resource languages pose a significant barrier to the application of advanced linguistic methodologies and technologies. To overcome this limitation, we present an innovative approach utilizing language-agnostic models, which are designed to be flexible and adaptable across different languages, enabling cross-linguistic knowledge transfer. A language-agnostic model projects sentences from multiple languages into a shared high-dimensional space, bringing closer the similar meanings. This approach helps overcome the challenges of low-resource languages and extends the reach of advanced language technologies. We selected LaBSE~\cite{LABSE}, a language-agnostic model based on BERT~\cite{BERT}, as our primary choice for encoding low-resource languages, due to its ability to learn how to map sentences from different languages into a high-dimensional shared space.  Notably, LaBSE differs from Multilingual BERT, which may not be as effective as a language-agnostic model for low-resource languages~\cite{no_multilingual_BERT}. We also considered LASER~\cite{LASER}, an LSTM~\cite{LSTM} model partially trained on Bambara, but its intrinsic bias and lack of transformer architecture led us to discard it after initial experiments. 


\subsubsection{Optimization of computational resources}
\label{subsub:opt_cr}
Optimization of computational resources increases the accessibility of NLP techniques for researchers and developers in resource-constrained environments by enabling experiments that do not need an expensive infrastructure. \\ Resource optimization lowers the risk of overfitting and improves model performance by enabling deep neural networks to be trained with fewer parameters, which is particularly challenging in situations with sparse data. Combining methods like transfer learning, using pre-trained models on languages with large amounts of resources, and model regularization can improve performance on various tasks while making better use of the available resources.
\paragraph{LoRA}
Low-Rank Adaptation (LoRA)~\cite{Hu2021} is selected for its efficiency and employed to address the above challenges in low-resource settings such as Bambara. LoRA enables model adaptation by introducing trainable low-rank matrices into Transformer layers. It freezes the original model weights and significantly reduces the number of trainable parameters. This approach allows effective fine-tuning without requiring large datasets or altering the inference latency. Its design promotes parameter reuse, supporting efficient task switching and making it a practical solution for extending NLP capabilities to under-resourced languages.
\subsubsection{Cyclical learning rate}
One of the most important parameters in training is the learning rate. The parameters (weights) \(\theta_t\) of a deep neural network are typically updated in several subsequent epochs, each at time \(t\), by gradient stochastic descent, by: $$\theta_{t+1} = \theta_{t} - \epsilon_t \frac{\partial L_t}{\partial \theta}$$
where \( L_t \)  is the loss function observed at time \(t\) and \( \epsilon_t \) is the learning rate that we change cyclically with \(t\). It is well established in literature that a too small learning rate hinders the training algorithm by slowing down its convergence, while a too large learning rate can cause the algorithm to diverge. \\ One approach to managing the learning rate is to use a cyclic schedule~\cite{Smith2017}: it is characterized by a linear function oscillating between a minimum and a maximum throughout training, and yields remarkable results with minimal computational cost.\\

\subsubsection{Transferring knowledge and learning through model distillation}
\label{subsub:t_learning}
Knowledge distillation~\cite{Thapa2022} is a machine learning approach that aims to enhance the performance of a smaller, "student" model by transferring knowledge from a larger, "teacher" model. This process entails training the student model to replicate the behaviour of the teacher model, thereby leveraging the teacher expertise to inform the student model capabilities. Recent advances in distillation have resulted in significant enhancements in performance across a range of tasks and benchmarks~\cite{mercer2025briefanalysisdeepseekr1}. 
However, the majority of existing research is focused on achieving state-of-the-art results for specific problems, with limited attention given to understanding the underlying process and its behaviour under different training conditions. Similarly, Transfer Learning is a powerful technique that facilitates the accelerated training of neural networks on small datasets by leveraging representations learned from a related, pre-existing problem.
Transfer learning applied to Machine Translation can be considered as a methodology that involves a parent model teaching a child model the specific parameters of a language pair. \\

\section{Data and sources}
\label{data-and-sources}

The data employed in our pipelines consists of aligned sentence pairs in Bambara, French, and English, sourced from a range of heterogeneous collections. Acquiring high-quality parallel corpora for Bambara is particularly challenging due to the predominantly oral tradition of the language. Although there has been substantial linguistic research on this language,~\cite{vydrin2016texts, tapo-etal-2025-bayelemabaga} very few data are suitable for training neural machine translation systems. For this study, four multilingual resources were identified as particularly valuable: The Dokotoro manual, the Bayelemagaba Dataset, the Mafand-MT corpus, the NLLB Corpus, the Bambara Lexicon.

\paragraph{The Dokotoro manual.} This manual is a multilingual adaptation of \emph{"Where There Is No Doctor"}, a widely used 600-page community health guide developed by the Hesperian Health Guides\footnote{\url{https://hesperian.org/}}. The Dokotoro Project translated and aligned the content of the manual in Bambara, French, and English. The aligned translations are available at the official website\footnote{\url{https://dokotoro.org/}}. From the Dokotoro manual, we retrieve a total of 9075 aligned sentences.

\paragraph{The Bayelemagaba Dataset.}
Bayelemagàba is a parallel French-Bambara dataset consisting of $46,976$ sentence pairs extracted from 264 text sources, including periodicals, books, short stories, blog posts, and religious texts such as the Bible and Quran. It is already split into training (80\%), validation (10\%), and test (10\%) sets and it is openly available on HuggingFace\footnote{\url{https://huggingface.co/datasets/RobotsMaliAI/bayelemabaga}}.

\paragraph{The Mafand-MT corpus.}
The Mafand-MT corpus\footnote{\url{https://huggingface.co/datasets/masakhane/mafand}} (Masakhane Anglo \& Franco Africa News Dataset for Machine Translation) is a high-quality parallel dataset for low-resource machine translation in the news domain. It includes professionally translated sentence pairs from English or French, spanning topics such as politics, education, culture, and religion. For Bambara-French, the dataset comprises $3,014$ training, $1,501$ validation, and $1,501$ test examples. 

\paragraph{The NLLB corpus.} The NLLB (No Language Left Behind) corpus\footnote{\url{https://opus.nlpl.eu/NLLB/fr&bm/v1/NLLB}} is a large multilingual dataset for machine translation, including separate Bambara–English and Bambara–French subsets, which are not mutually aligned. We selected the Bambara–French dataset, which contains $307,529$ sentence pairs mined by Meta AI. Although extensive, the corpus includes noise and limited high-quality alignments, requiring preprocessing to improve sentence consistency. 

\paragraph{The Bambara Lexicon.}
The Bambara Lexicon\footnote{\url{http://www.bambara.org/en/index.htm}} is a compact yet valuable multilingual dictionary. It includes domain-specific terms and definitions in areas such as agriculture, weather, geography, and related fields, presented in Bambara, French, and English. The lexicon offers aligned terms with definitions and example usages. From this resource, we extracted $1,394$ aligned terms and 230 aligned example sentences.

\subsection{Data Preprocessing}
To prepare a multilingual dataset for neural machine translation, we used all sentences from the specified sources, excluding the Bayelemagaba and Mafand-MT test sets, treating the result as a benchmark. We then applied a preprocessing pipeline to enhance data quality and alignment. Bambara–French sentence pairs were cleaned, standardized, and aligned semantically. The main preprocessing steps are outlined below:

\begin{itemize}
     \item \textbf{Text normalization:} Regular expressions and custom rules were applied to fix punctuation, remove extra spaces, and eliminate unwanted characters (e.g., ``<'', ``\{'', ``»'') or enumerations (e.g., ``1)'', ``a)'') when they appeared only in one of the aligned sentences.

    \item \textbf{Remove HTTP links:} Rows containing hyperlinks were removed entirely to exclude web addresses and metadata that could degrade translation quality.

    \item \textbf{Anomalous character removal:} Informal or noisy sequences with exaggerated character repetition (e.g., \texttt{``loooool''}, \texttt{``nooooooo''}) were identified and removed using pattern matching, as these are often informal or noisy data.
    
    \item \textbf{Emoji filtering:} Emojis that appeared on only one side of a sentence pair were removed to maintain semantic symmetry.
    
    \item \textbf{Remove duplicates:} After normalization, duplicate entries were removed to prevent bias during model training and evaluation.
    
\end{itemize}
\subsection{The novel Yiri Dataset}
Following the data preprocessing phase, we constructed a parallel corpus referred to as the \textit{Yiri} dataset. The dataset was divided into three subsets: 80\% for training, 10\% for validation, and 10\% for testing and contains a total of $353,629$ aligned Bambara-French sentences. Table~\ref{tab:yiri} summarizes the distribution of sentence counts and total word tokens across the three subsets.

\begin{table}[h]
\caption{Distribution of tokens and sentences in the three partitions (train, validation and test sets) of the Yiri dataset\label{tab:yiri}}
\centering
\vspace{0.5cm}
\begin{tabular}{lrrr}
\hline
\textbf{Subset} & \textbf{Sentences} & \textbf{Bambara Tokens} & \textbf{French Tokens} \\ \hline
Training    	& 282,903        	& 3,319,156          	& 2,943,235         	\\
Validation  	& 35,363         	& 413,954            	& 366,705           	\\
Test        	& 35,363         	& 416,760            	& 369,415           	\\
\textbf{Total}  & \textbf{353,629}   & \textbf{4,149,870} 	& \textbf{3,679,349}	\\ \hline
\end{tabular}
\end{table}

\subsection{Benchmarks}
\label{benchmarks}
As previously noted, the Yiri dataset excludes the Bayelemagaba and Mafand-MT test sets, which are used in the literature as benchmarks for evaluating machine translation models involving Bambara. We assess our models on the following three external benchmarks, as well as on the Yiri test set: Bayelemagaba test set, Mafand-MT test set, FLORES+ devtest split.\\
FLORES+~\cite{nllb-24} is a multilingual evaluation dataset developed by the Open Language Data Initiative. It includes high-quality translations of $1,012$ devtest sentences from English into over 200 language varieties, including French and Bambara.

\section{The first pipeline Transformer-based: Transformer From Scratch}
\label{The Transformer-based pipeline}
We conduct initial experiments in machine translation using a basic Transformer~\cite{vaswani2023attentionneed}.
The pipeline is shown in Figure~\ref{fig:first_second_pipeline}.
We used as a benchmark the Yiri dataset as a baseline for comparison with the other two methods. Our implementation is based on the JoeyNMT framework\footnote{\url{https://github.com/joeynmt/joeynmt}}, which supports reproducible experimentation via YAML configuration files. Following Tapo et al.~\cite{tapo-etal-2020-neural}, we explore various architectural and training hyperparameters to optimize configurations for low-resource translation tasks. Performance is evaluated using BLEU~\cite{papineni2002bleu} and chrF~\cite{popovic2015chrf} metrics.
\subsection{Model Architecture and Training Configuration}
We adopt a Transformer architecture with three fixed configurations tailored to the limited size of the training dataset. These variants balance model capacity and computational efficiency, as detailed in Table~\ref{tab:architecture}: Transformer 1 uses 4 layers, 4 attention heads, a hidden size of 128, and a feed-forward size of 512; Transformer 2 increases these to 6 layers and 256/1024 dimensions, while Transformer 3 further expands the hidden and feed-forward sizes to 512 and 2048, respectively. All models use a dropout of $0.2$, tied softmax weights, Xavier initialization, and the Adam optimizer ($\beta_1 = 0.9$, $\beta_2 = 0.999$). Training runs for up to 300 epochs with early stopping after four stagnant epochs. Beam search (widths of 5 and 10) is used for decoding. The hyper-parameters settings are evaluated on the Yiri validation set during training, using BLEU and chrF. Results from experiments on these configurations are reported in Section~\ref{experiments-and-results}.

\begin{table}[h]
\caption{Parameters and values explored during training of the first pipeline
\label{tab:architecture}}
\centering
\vspace{0.5cm}
\normalsize
\begin{tabular}{@{}ll@{}}
\toprule
\textbf{Parameter}                  & \textbf{Values Explored}     \\ \midrule
Model architecture                  & Transformer 1, 2, 3          \\
Embedding dimension                 & 128, 256, 512                \\
Number of epochs                    & 100, 200, 300                \\
Token batch size                    & 4096, 8192, 16384            \\
Beam search width                   & 5, 10                        \\
Learning rate (initial)             & $1e^{-5}$, $2e^{-5}$, $5e^{-5}$              \\
Minimum learning rate               & $1e^{-10}$                        \\
Learning rate decrease factor       & 0.5                          \\
Dropout                             & 0.2, 0.3                     \\
\bottomrule
\end{tabular}
\end{table}

\subsection{Segmentation}
\label{segmentation}
Due to the lack of standardized tokenizers for Bambara and the small size of the Yiri dataset, we apply subword segmentation using Byte Pair Encoding (BPE) with a shared vocabulary for French and Bambara. Implemented via the \texttt{subword-nmt} toolkit\footnote{\url{https://pypi.org/project/subword-nmt/}}, BPE is trained only on the Yiri training set to avoid information leakage. Merge size was tuned in JoeyNMT experiments, with 5000 merges selected based on validation performance and vocabulary coverage. This approach reduces out-of-vocabulary issues by breaking rare words into frequent subword units, crucial for low-resource translation.

\section{The second pipeline: Instructor-based LLama3}
\label{llama}
\begin{figure}[hbt]
  \centering
  \includegraphics[width=\linewidth]{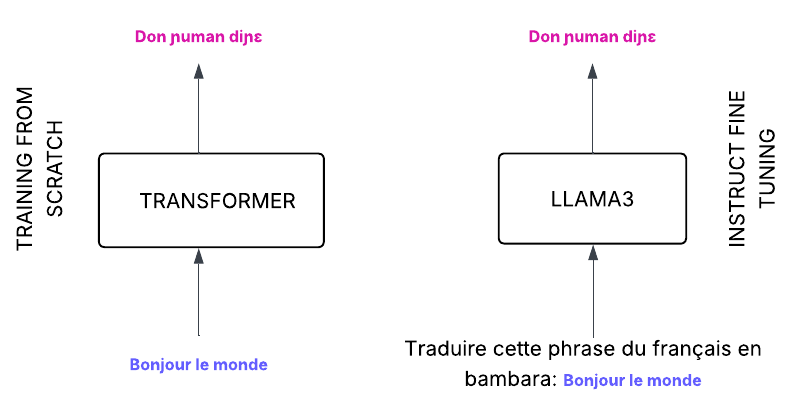}
  \caption{Differences and similarities of the first two pipelines of Section~\ref{The Transformer-based pipeline} and Section~\ref{llama} for training automatic translation\label{fig:first_second_pipeline}}
  \ \\[0.3em]
\end{figure}

The second pipeline is shown in Figure~\ref{fig:first_second_pipeline}.
It considers instruction tuning as a strategy to adapt LLMs to translation tasks in under-resourced settings, even with limited supervision. Specifically, we investigated the use of instruction-tuned versions of LLaMA 3 for the French–Bambara translation, leveraging small-scale parallel corpora to teach the model to follow translation prompts in an interactive format.
We fine-tuned both the 3B and 8B variants of LLaMA 3 targeting machine translation from French to Bambara. For training, we used the Yiri dataset. 
Following the preprocessing steps described in Section~\ref{segmentation} we adapted the datasets for instruction tuning by converting the parallel sentence pairs into instruction-based prompts. Each instance was structured as follows: the system prompt was fixed as \textit{"Traduire cette phrase du français en bambara"}, the user input consisted of the French sentence, and the assistant output corresponded to the expected translation in Bambara. This format aligns with instruction-following paradigms to enable the model to learn the task within a natural language interaction framework.

\subsection{Setting}
For the fine-tuning experiments, we explored a range of hyperparameter configurations to assess their impact on translation quality. Specifically, we varied the learning rate among three values: $5e^{-5}$, $2e^{-5}$, and $1e^{-5}$. The models were fine-tuned for 3, 5, or 10 epochs, and we tested two learning rate schedulers: linear and constant.
We employed LoRA to reduce the number of trainable parameters during fine-tuning. We tested rank values of 8 and 16, with the corresponding LoRA scaling factor \(\alpha\) always set to twice the rank (i.e., $alpha = 2r$), as commonly recommended in the literature.
Due to memory constraints, we used a batch size of 2 for LLaMA 3 8B and 8 for LLaMA 3 3B.

\section{LoReB: The third pipeline with BERT-enhancement}
\label{BERT-enhancement-pipeline}
We introduce \textbf{LoReB}, a fine-tuning pipeline for LLMs designed to address challenges in LRLs. It integrates cross-lingual distillation, parameter-efficient tuning via LoRA, and lightweight BERT-based decoding enhancement. While these techniques have been used individually in high-resource settings, their application to low-resource languages, such as Bambara, has not been explored in this unified way.\\Cross-lingual distillation has shown strong results in multilingual contexts but typically requires full model fine-tuning with substantial resources. To make this feasible in low-resource settings, we employ AdamW and LoRA. AdamW is used because it is known for its resistance to gradient instability that often arises when fine-tuning on limited data. LoRA is employed as it is considered best practice in parameter-efficient fine-tuning (PEFT). By freezing parts of the model, LoRA reduces the number of trainable parameters while still allowing effective learning. This approach helps mitigate the risk of overfitting, enabling us to preserve the core model knowledge.\\
BERT-based decoding enhancements are employed in LLR for the first time to improve fluency and semantic alignment.\\
The pipeline focuses primarily on improving the \emph{encoding} phase through multilingual knowledge transfer. By applying our distillation strategy to LaBSE, an agnostic BERT model, we enable it to internalize robust representations of LRLs. As demonstrated in our experimental results, this encoding strategy alone yields promising performance improvements over baseline approaches, particularly in the case of Bambara.
Additionally, LoReB incorporates a lightweight BERT-based enhancement during decoding, with the T5 decoder and a layer before it, as shown in Figure~\ref{fig:loreb_enh}, to address the dimensionality mismatches between the embeddings. 
Our evaluation shows that this decoding refinement contributes to improved fluency and consistency, even in resource-constrained scenarios.
Together, these components form a coherent pipeline that offers a reusable foundation for future work on LRLs.

\subsection{Applying Multi-language Distillation}
Of paramount importance for LoReB is the application of language distillation during fine-tuning, introduced in Section~\ref{lit_rev}. In the implementation of the pipeline for fine-tuning LoReB, the Formula \ref{eq:mse_loss} is used as loss during training. Shown as $TM(s_j)$ is the \emph{j}-th
embedding created from the teacher model $TM$ to the \emph{j}-th example of the source language $s_j$, while $SM(s_j)$ and $SM(t_j)$ are the \emph{j}-th embeddings respectively in the \emph{j}-th examples from the source and target languages $s_j$ and $t_j$ represented by the student model $SM$. $|B|$ is the number of sentences in the mini-batch $B$ of the training loop.
\begin{equation}
\label{eq:mse_loss}
\text{ $L_t$ = $\frac{1}{|B|} \sum_{j \in B} \left( TM(s_j) - SM(s_j) \right)^2 + \left( TM(s_j) - SM(t_j) \right)^2 $}
\end{equation}
\subsection{Applying resources optimization techniques}
We conducted preliminary experiments comparing the out-of-the-box LaBSE model with our fine-tuned version. In its final pipeline version, LoReB employs the AdamW optimizer ~\cite{Loshchilov2019}, which is more resilient to vanishing gradient issues than Adam optimizer ~\cite{Kingma2017}.
A generic class was developed to support both encoder-only and encoder-decoder LLM architectures, incorporating the LoRA optimization technique.
\begin{figure}[hb]
  \centering
  \includegraphics[width=\linewidth]{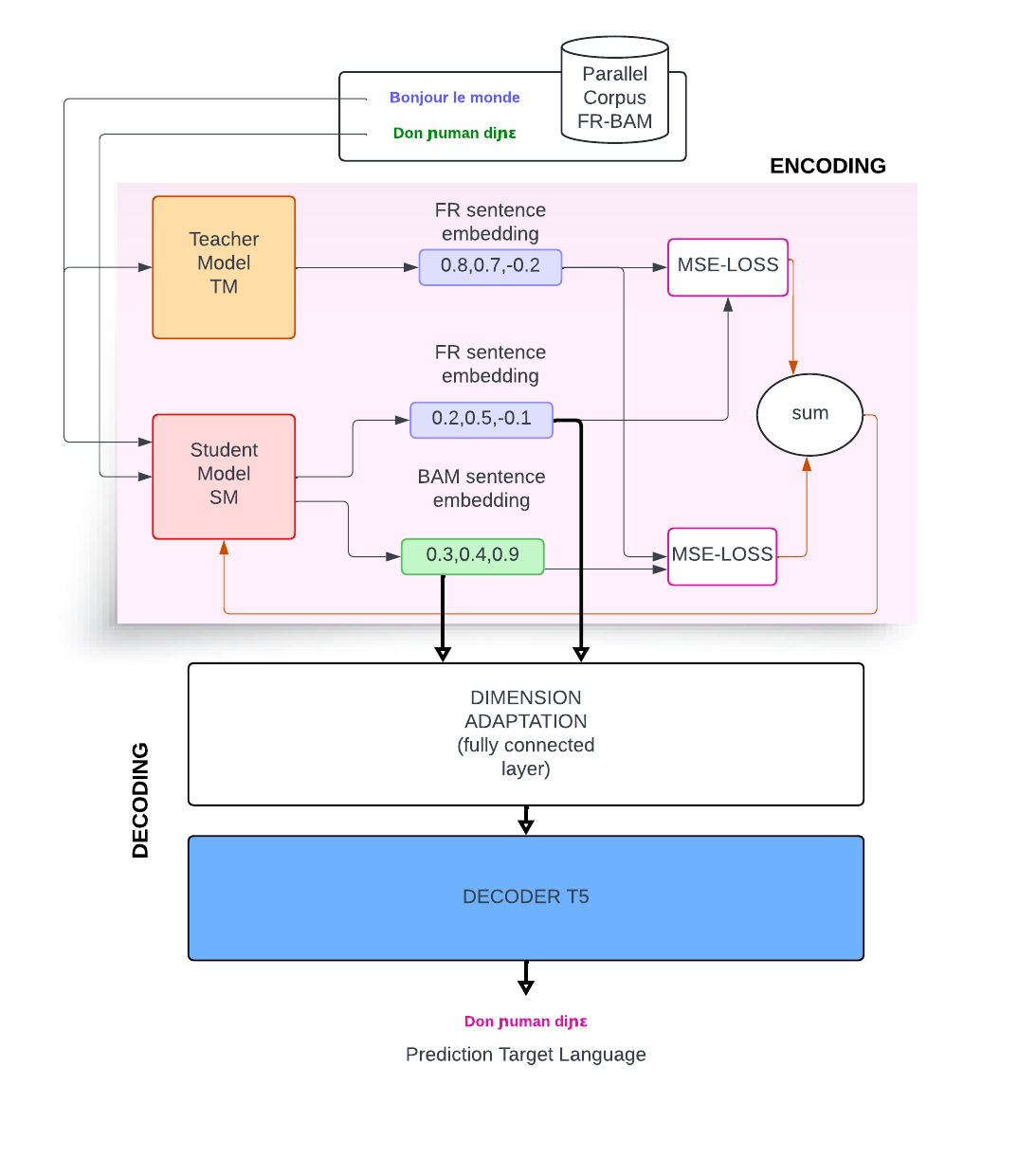}
  \caption{LoReB pipeline: The upper section is the distillation, while the lower section is the decoding with a fully connected layer to adapt dimensions
  \label{fig:whole_loreb}}
  \ \\[0.3em]
\end{figure}

\section{Experiments and results}
\label{experiments-and-results}
This section presents the results of all experiments conducted using the various pipelines. 

\subsection{First pipeline on Transformer From Scratch}
Table~\ref{tab:transformer-results} shows BLEU and chrF scores for three alternative Transformer configurations across various test sets. Transformer 2 (T2) consistently outperformed Transformer 1 (T1) and Transformer 3 (T3), demonstrating the best balance between capacity and generalization in low-resource settings. On the Yiri test set, T2 achieved the highest scores (BLEU \textbf{33.81\%}, chrF \textbf{41.00\%}), exceeding T1 by 6.77\% BLEU and T3 by 8.19\%. Similar trends were observed in the Bayelemagaba (10.28\% BLEU, 21.01\% chrF), MAFAND-MT (9.44\% BLEU, 20.12\% chrF), and FLORES+ (7.63\% BLEU, 18.07\% chrF) sets. These results highlight that smaller, well-balanced models like T2 can outperform larger ones by generalizing better and avoiding overfitting in low-resource scenarios.

\begin{table}[]
\caption{BLEU and chrF scores for Transformer configurations\label{tab:transformer-results}}
\centering
\vspace{0.5cm}
\setlength{\tabcolsep}{1.5pt}
\begin{tabular}{@{}ccccccccc@{}}
\toprule
\textbf{Benchmarks} &
  \textbf{Model} &
  \textbf{\begin{tabular}[c]{@{}c@{}}Beam \\ Width\end{tabular}} &
  \textbf{\begin{tabular}[c]{@{}c@{}}Train\\ Epochs\end{tabular}} &
  \textbf{\begin{tabular}[c]{@{}c@{}}Batch\\ size\end{tabular}} &
  \textbf{Dropout} &
  \textbf{LR} &
  \makecell{\textbf{BLEU} \\ \textbf{[\%]}}
 &
  \textbf{chrF} \\ \midrule
Yiri test set & T2 & 10 & 300 & 16384 & 0.2 & $5e^{-5}$ & \textbf{33.81} & \textbf{41.00} \\
                               & T1 & 5  & 200 & 8192  & 0.3 & $5e^{-5}$ & 27.04          & 34.03          \\
                               & T3 & 5  & 300 & 8192  & 0.3 & $5e^{-5}$ & 25.62          & 32.12          \\ \midrule
Bayelemagaba  & T2 & 10 & 300 & 16384 & 0.2 & $5e^{-5}$ & \textbf{10.28} & \textbf{21.01} \\
                               & T1 & 5  & 200 & 8192  & 0.3 & $5e^{-5}$ & 7.86           & 17.43          \\
                               & T3 & 10  & 300 & 8192  & 0.3 & $5e^{-5}$ & 6.43           & 16.11          \\ \midrule
MAFAND-MT     & T2 & 10 & 300 & 16384 & 0.2 & $5e^{-5}$ & \textbf{9.44}  & \textbf{20.12} \\
                               & T1 & 10  & 200 & 8192  & 0.3 & $5e^{-5}$ & 6.93           & 15.02          \\
                               & T3 & 5  & 300 & 8192  & 0.3 &$ 5e^{-5}$ & 6.01           & 14.62          \\ \midrule
FLORES+        & T2 & 10 & 300 & 16384 & 0.2 & $5e^{-5}$ & \textbf{7.63}  & \textbf{18.07} \\
                               & T1 & 5  & 200 & 8192  & 0.3 & $5e^{-5}$ & 5.01           & 14.52          \\
                               & T3 & 5  & 300 & 8192  & 0.3 & $5e^{-5}$ & 4.26           & 13.16          \\ \bottomrule
\end{tabular}
\end{table}
Despite the observed performance, T2’s results remain a bit below those reported in the literature. On MAFAND-MT, Adelani et al.~\cite{adelani2022mafind} report $22.7$\% BLEU and $48.2$ chrF. For Bayelemagaba, Tapo et al.~\cite{tapo-etal-2025-bayelemabaga} achieve $35.7$\% BLEU and $25.8$ chrF. On FLORES+ (fra–bam), the NLLB Team~\cite{nllbteam2022nllb} report $24.7$\% BLEU and $49.9$ chrF. 

\subsection{Second pipeline: Instructor-based LLama3}
Models were evaluated over 10 translation runs on various test sets whose results are in Table~\ref{tab:translation-results}. On the FLORES+ benchmark, the LLaMA 3 8B model achieved an average BLEU score of $2.8\%$ and a chrF of $15.7$. The smaller LLaMA 3 3B model scored $1.4\%$ BLEU and $10.2$ chrF. On the MAFAND-MT test set, LLaMA 3 8B outperformed the 3B model with a BLEU score of $8.2\%$ and a chrF of $28.3$, while the 3B model scored $6.3\%$ BLEU and $23.5$ chrF. The LLaMA 3 8B model also led in the Bayelemabaga test set with a BLEU of $9.82\%$ and chrF of $19.00$, while the 3B model scored $3.00\%$ BLEU and $11.5$ chrF. The results highlight the significant impact of the size of the model on translation quality. Domain adaptation through fine-tuning improved performance, as seen in higher scores on the MAFAND test set compared to the zero-shot FLORES+ evaluation. Although scores were low, they showed substantial improvement over non-finetuned baselines, which consistently achieved BLEU scores around $2\%$.

\begin{table}[]
\caption{Translation performance (BLEU and chrF) of fine-tuned LLaMA 3 on French-to-Bambara. Results are averaged over 10 runs
\label{tab:translation-results}}
\centering
\vspace{0.5cm}
\begin{tabular}{@{}cccc@{}}

\toprule
\textbf{Model} & \textbf{Dataset} & \textbf{BLEU [\%]} & \textbf{chrF} \\ \midrule
LLaMA 3 8B     & FLORES+         & 2.8           & 15.7          \\
LLaMA 3 3B     & FLORES+         & 1.4           & 10.2          \\
LLaMA 3 8B     & MAFAND-MT        & 8.2           & 28.3          \\
LLaMA 3 3B     & MAFAND-MT        & 6.3           & 23.5          \\
LLaMA 3 8B     & Bayelemabaga     & 9.82          & 19.00         \\
LLaMA 3 3B     & Bayelemabaga     & 3.00          & 11.5          \\ \bottomrule
\end{tabular}
\end{table}

\subsection{The third pipeline with LoReB Distillation}
\label{distillation}
We show the experimental results of the application of the encoder of LoReB pipeline.
It was fine-tuned on approximately $110,000$ samples, half of which were pairs of English-Bambara and the other half of French-Bambara. The datasets used were Dokotoro and a randomly generated subset of sentences from NLLB. The encoder was fine-tuned for 100 epochs.
\subsubsection{Feature extraction: comparison between models}
\begin{figure}[htb]
\centering
    \includegraphics[width=0.9\linewidth]{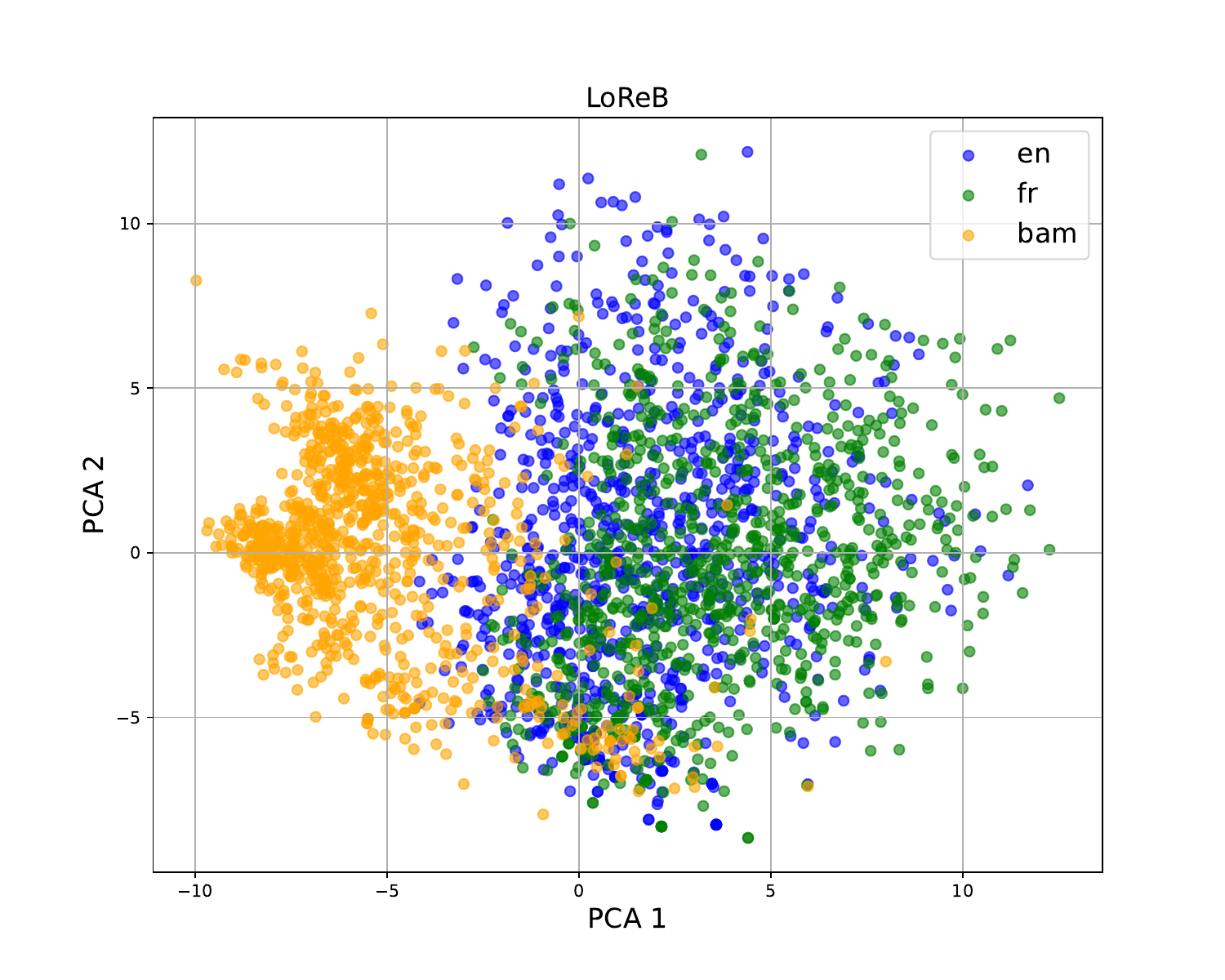}
\caption{LaBSE embeddings after LoReB fine-tuning
\label{fig:pca_combined}}
\vspace{0.5cm}
\ \\[0.1em] 
\end{figure}

The fine-tuning process of the LaBSE model involves different stages and settings, each characterized by varying degrees of model adaptation and performance using only the embeddings. For instance, Figure~\ref{fig:pca_combined} shows the embedding space (originally with 768 dimensions) of Dokotoro sentences in English, French and Bambara.  The figure shows the result of a PCA decomposition of the embeddings, projected along the first two principal components, only. Since the projection is an approximation, the picture could reveal some kind of disalignment. However, as shown in Figure~\ref{fig:pca_combined} high-resource languages (blue and green) cluster more closely than the low-resource language (yellow), indicating stronger semantic alignment for high-resource languages. After the training epochs, however, the cluster of Bambara embeddings shows the clear tendency to converge into a more unified embedding space.

\subsubsection{Distributions of distances}
\begin{figure}[htb]
\centering
\includegraphics[width=0.9\linewidth]{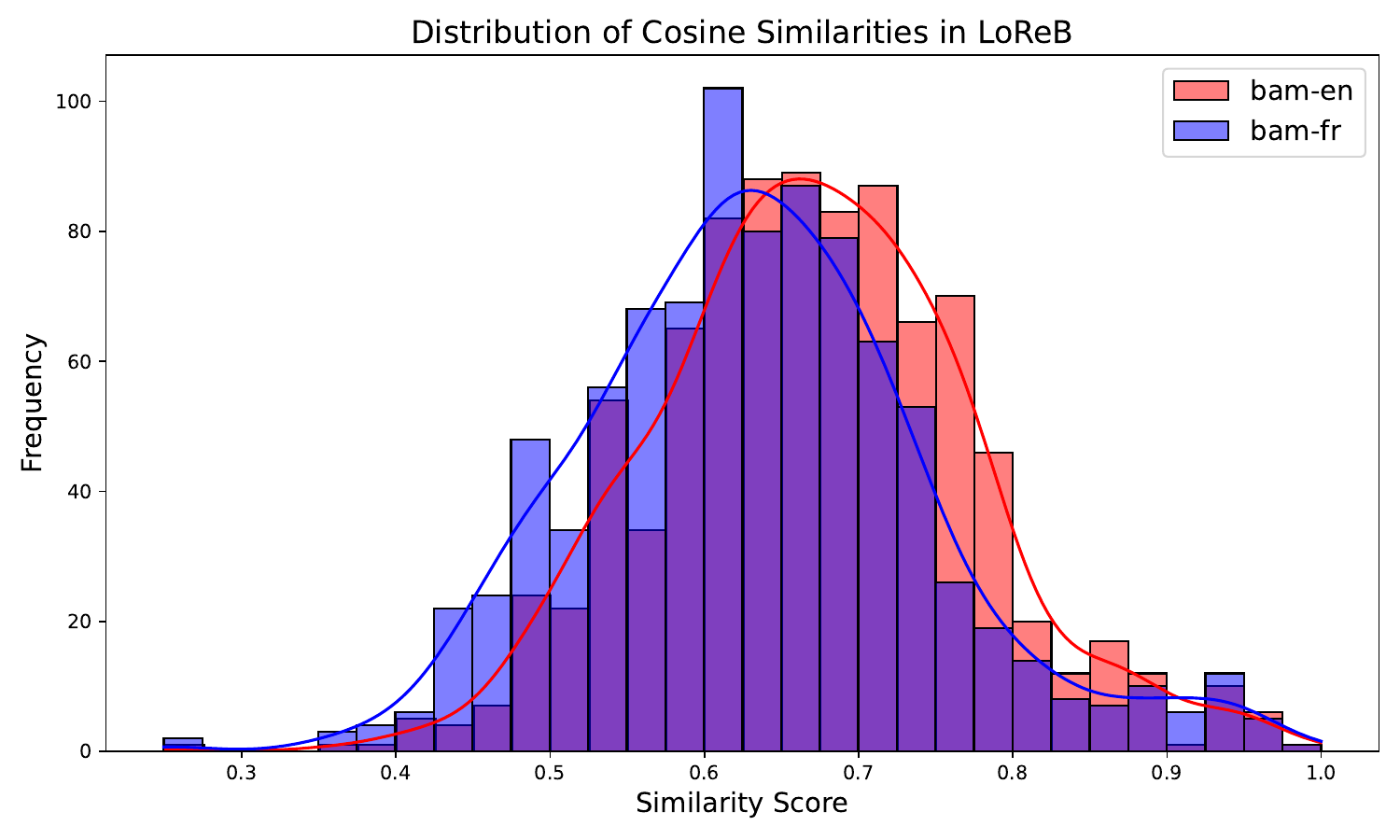} 
\caption{Distribution of cosine similarity on the embeddings of source and target sentences produced by the encoder in LoReB\label{fig:cosine_similarity}}
\ \\[0.3em]
\end{figure}

Figure~\ref{fig:cosine_similarity} shows the distribution of cosine similarity scores between the embeddings of the source language and the embeddings of the target language after the encoding phase of LoReB. It has a roughly normal shape, centred around $0.65-0.7$. This indicates that most sentence pairs embeddings are moderately to highly similar, suggesting a generally strong semantic alignment. A smaller number of pairs show very high or very low similarity, reflecting occasional near-duplicates or mismatches. Overall, the distribution supports the presence of meaningful correspondence between the encoded sentences.

\subsubsection{LoReB translation}
\begin{figure}[hbt]
  \centering
  \includegraphics[width=1\linewidth]{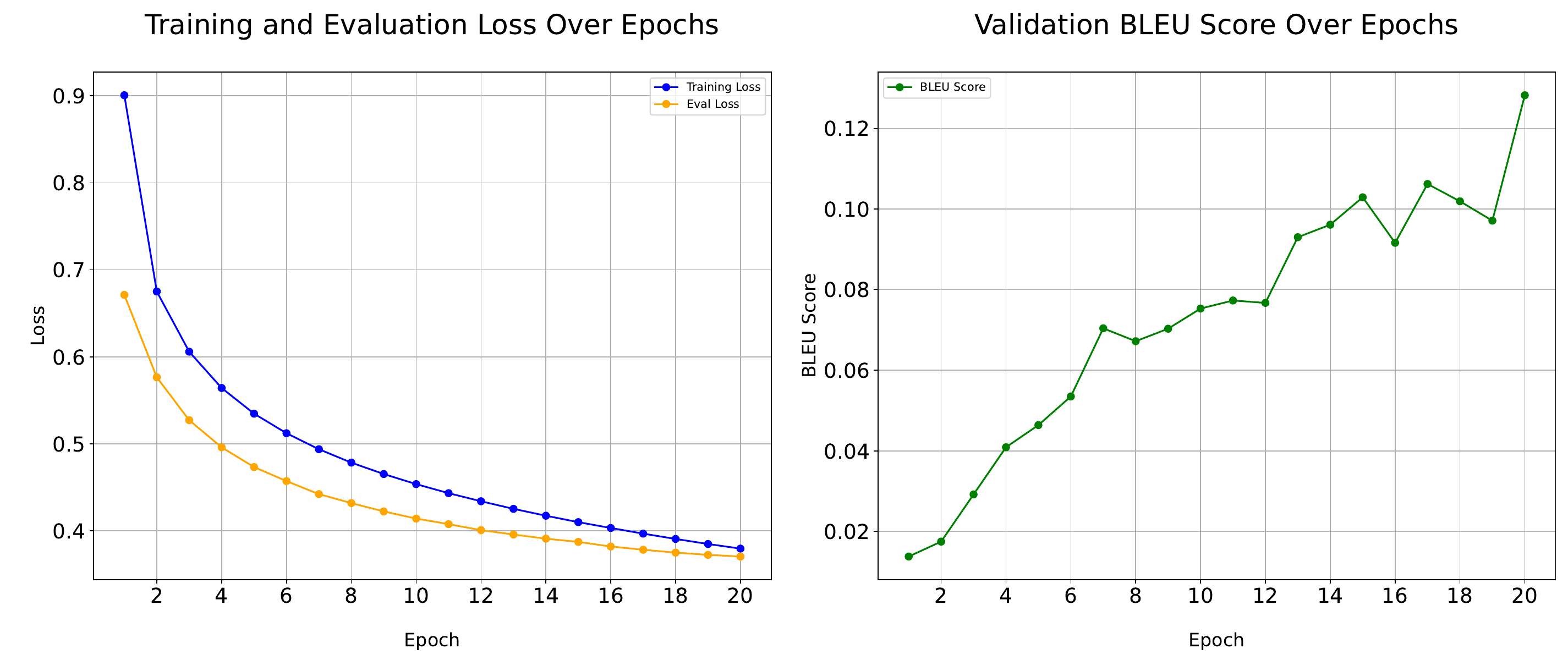}
  \caption{The training progress in LoReB: (Left) the convergence in loss in training and evaluation sets; (Right) the increase in average BLEU score
  \label{fig:loreb_enh}}
  \ \\[0.3em]
\end{figure}

Figure~\ref{fig:loreb_enh} shows the progress of training of the decoder T5 needed to generate a translation from a source sentence (French) into a target one (Bambara).
Figure~\ref{fig:loreb_enh} shows a consistent decrease in average loss and a corresponding increase in BLEU score over epochs, indicating effective learning and improved alignment with the reference targets. 

The experimental results of LoReB in the translation from French to Bambara on the benchmarks test sets are shown in Table~\ref{tab:translation-results-T5}.

\begin{table}[]
\caption{Translation performance (BLEU and chrF) of T5 in LoReB on French-to-Bambara\label{tab:translation-results-T5}}
\centering
\vspace{0.5cm}
\begin{tabular}{ccc}
\hline
\textbf{Dataset} & \textbf{BLEU {[}\%{]}} & \textbf{chrF} \\ \hline
Yiri test set    & 13.21                  & 34.15         \\
Bayelemabaga     & 2.6                    & 27.39         \\
MAFAND-MT        & 1.12                   & 33.16         \\
FLORES+          & 1.20                   & 28.17         \\ \hline
\end{tabular}
\end{table}

From the experimental results of the BLEU and chrF scores of the three pipelines, it seems that FLORES+ allows less promising results than the other datasets. Similar results have also been observed in other works on NMT for LRLs~\cite{silva-etal-2024-benchmarking} (although not with  Bambara and on FLORES-200, which is similar enough to FLORES+ but simpler, because it focuses on less nuanced contexts).

Finally in Table~\ref{tab:computational-resources} we compare the amount of computational resources employed for the three pipelines. For time and resource constraints we had to use different HPC architectures in which T4 support for training is limited, L4 discrete and A100 excellent. 
\begin{table}[]
\caption{Amount of computational resources in training\label{tab:computational-resources}}
\centering
\vspace{0.5cm}
\begin{tabular}{ccccc}
\hline
\textbf{Pipeline}          & GPU type & VRAM   & Epochs & Time (hrs)\\ \hline
Transformer from Scratch   &  L4      &  24 GB &  300   &   24 \\
Instructor-based LLama3    &  A100    &  40 GB &  10    &    4 \\
LoReB  (encoding)          &  T4      &  16 GB &  100   &  200 \\
LoReB  (decoding)          &  A100    &  40 GB &  20    &   10 \\ \hline
\end{tabular}
\end{table}

\section{Conclusions and future work}
This study presents three NMT pipelines specifically designed for LRL settings, focused on Bambara, a Mandè language. The proposed approaches include: training a Transformer model from scratch, fine-tuning a LLaMA 3 model using instruction-based parameter-efficient tuning, and implementing a hybrid system that integrates LaBSE embeddings with a T5 model for sequence generation. Despite its relative simplicity, the analysis shows that the Transformer model used in the initial pipeline consistently achieves the best performance across all benchmarks in this study. In particular, T2 outperforms both the LLaMA 3 3B and 8B models, as shown in Tables~\ref{tab:transformer-results} and \ref{tab:translation-results}. Even when compared to the best-performing LLaMA 3 8B configurations, T2 delivers significantly stronger results, especially on MAFAND-MT and Bayelemagaba benchmarks, where both BLEU and chrF scores are noticeably higher. Regarding the LoReB pipeline results, compared to the other two are lower, but promising. As a whole, these findings show that in low-resource settings, such as our translation task with a dataset of approximately $300,000$ aligned sentence pairs, simpler architectures like the Transformer can yield good performance.  This is likely because larger models like LLAMA 3 may require significantly more data and computational resources to reach their full potential, even using parameter-efficient fine-tuning, often leading to underperformance in constrained environments.
Future work aims at integrating human evaluation that will complement automatic metrics such as BLEU and chrF, offering a more nuanced assessment of translation quality. These directions aim to support broader efforts toward equitable access to language technologies for marginalized linguistic communities.

\label{decoder-and-translation}

\bibliography{ecai-sample-and-instructions}

\end{document}